\documentclass[runningheads]{llncs}
\usepackage{times} 
\usepackage{graphicx}
\usepackage{latexsym}
\usepackage{algorithm}
\usepackage[noend]{algpseudocode}
\usepackage{amssymb}
\usepackage{color}

\newcommand{\tip}{{\bf T}}
\newcommand{\alc}{\mathcal{ALC}}
\newcommand{\alct}{\mathcal{ALC}+\tip_{\bf R}}
\newcommand{\TBox}{{\mathcal{T}}}
\newcommand{\ABox}{{\mathcal{A}}}
\newcommand {\allworlds} {\mathcal{W}_{\mathcal{K}}}
\newcommand {\kk} {\mathcal{K}}
\newcommand{\RR}{{\mathcal{R}}}
\newcommand{\TT}{{\mathcal{T}}}
\newcommand{\AAA}{{\mathcal{A}}}
\newcommand{\pfl}{\tip^{\textsf{\tiny CL}}}
\newcommand{\II}{{\mathcal{I}}}
\newcommand{\emme}{{\mathcal{M}}}
\newcommand{\PP}{{\mathbb{P}}}
\newcommand{\miaprob}{\alct^{\mathit{\textsf{\scriptsize P}}}}

\newcommand {\ent} {\mathrel{{\scriptstyle\mid\!\sim}}}

\newcommand {\nott} {\lnot}

\newcommand {\esse} {\mathbb{S}}
\newcommand {\CC} {\mathbb{C}}

\newcommand {\sx} {\langle}
\newcommand {\dx} {\rangle}

\newcommand {\unione} {\cup}

\newcommand {\tc} {\mid}

\newcommand {\diverso} {\neq}

\newcommand{\alctr}{\mathcal{ALC}+\tip_{\bf R}}
\newcommand{\alctmin}{\mathcal{ALC}+\tip_{{\bf R}}^{\mbox{\textsc{\scriptsize{\em RaCl}}}}}
\newcommand{\alctm}{\mathcal{ALC}+\tip_{{\bf R}}^{\mbox{\textsc{\scriptsize{\em RaCl}}}}}

\newcommand {\ellet} {\mathcal{L}_{\bf T}}

\newcommand{\be}{\begin{enumerate}}
\newcommand{\ee}{\end{enumerate}}

\newcommand{\hide}[1]{}

 \hide{
{\qed \end{trivlist}}

\newenvironment{definition}
{\begin{defi} \rm}{\qed \end{defi}}

\newenvironment {proofof}[2]
{\begin{trivlist} \item[] {\bf Proof of #1~\protect{\ref{#2}}.}}%
{\qed \end{trivlist}}

}

\newcommand {\pgrande} {\mathbb{P}}

\def \cases{\left \{\begin{array}{l}}
\def \endcases{\end{array}\right .}

\newcommand {\Pe} {{\bf P}}
\newcommand {\Ra} {{\bf R}}

\newcommand {\bes} {\begin{description}}
\newcommand{\ens} {\end{description}}

\newcommand {\beq} {\begin{quote}}
\newcommand {\enq} {\end{quote}}
\newcommand {\bit} {\begin{itemize}}
\newcommand {\enit} {\end{itemize}}

\newcommand{\sqset}{\sqsubseteq}

\newcommand{\mint}{\sqcap}

\newenvironment{pozz}{\color{black}}{\color{black}}

\begin{document}

\title{Reasoning about Typicality and Probabilities in Preferential Description Logics\vspace{-0.3cm}}

\author{Laura Giordano$^{\scriptsize \mbox{a}}$ \and Valentina Gliozzi$^{\scriptsize \mbox{b}}$ \and Antonio Lieto$^{\scriptsize \mbox{c}}$ \and \\ Nicola Olivetti$^{\scriptsize \mbox{d}}$ \and Gian Luca Pozzato$^{\scriptsize \mbox{e}}$}
\authorrunning{L. Giordano, V. Gliozzi, A. Lieto, N. Olivetti, G.L. Pozzato}
\titlerunning{Reasoning about Typicality and Probabilities in Preferential DLs}

\institute{$^{\mbox{a}}$ Dipartimento di Scienze e Innovazione Tecnologica - Istituto di Informatica,
Universit\`a  del Piemonte Orientale, Italy, \email{laura.giordano@uniupo.it} \\
$^{\mbox{b}}$ Dipartimento di Informatica,
Universit\`a di Torino, Italy, \email{valentina.gliozzi@unito.it}\\
$^{\mbox{c}}$ Dipartimento di Informatica,
Universit\`a di Torino, and ICAR-CNR (Palermo), Italy, \email{antonio.lieto@unito.it}\\
$^{\mbox{d}}$ Aix-Marseille University, France, \email{nicola.olivetti@univ-amu.fr}\\
$^{\mbox{e}}$ Dipartimento di Informatica,
Universit\`a di Torino, Italy, \email{gianluca.pozzato@unito.it}
}

\maketitle
\bibliographystyle{splncs}

\begin{abstract}
\vspace{-0.7cm}
In this work we describe preferential Description Logics of typicality, a nonmonotonic extension of standard Description Logics by means of a typicality operator $\tip$ allowing to extend a knowledge base with inclusions of the form $\tip(C) \sqsubseteq D$, whose intuitive meaning is that ``normally/typically $C$s are also $D$s''. This extension is based on a minimal model semantics corresponding to a notion of rational closure, built upon preferential models. We recall the basic concepts underlying preferential Description Logics. We also present two extensions of the preferential semantics: on the one hand, we consider probabilistic extensions, based on a distributed semantics that is suitable for tackling the problem of commonsense concept combination, on the other hand,
we consider other strengthening of the rational closure semantics and construction to avoid the so called ``blocking of property inheritance problem".
\end{abstract}

\section{Introduction}
\vspace{-0.3cm}
The family of Description Logics (for short: DLs) is one of the
most important formalisms of knowledge representation. They have a
well-defined semantics based on first-order logic and offer a good
trade-off between expressivity and complexity. DLs have been
successfully implemented by  a range of systems and they are at
the basis of languages for the semantic web such as OWL.

A  DL knowledge base (KB) comprises two components: the TBox,
containing the definition of concepts (and possibly roles) and a
specification of inclusion relations among them,  and  the ABox
containing instances of concepts and roles. Since the very
objective of the TBox is to build a taxonomy of concepts, the need
of representing prototypical properties and of reasoning about
defeasible inheritance of such properties naturally arises. The
traditional approach is to handle defeasible inheritance by
integrating some kind of nonmonotonic reasoning mechanism.  \color{black} This
has led to study nonmonotonic extensions of DLs
\cite{baader95a,baader95b,bonattilutz,donini98,donini2002,eiter2004,straccia93,casinistraccia2010}.
However, it is far from obvious to find a suitable  nonmonotonic extension for
inheritance with exceptions, according to the following desiderata: 1) The (nonmonotonic) extension must have a clear semantics and should be based on the same semantics as the underlying monotonic DL.
2) The extension should allow to specify prototypical properties in a natural and direct way.
3) The extension must be decidable, if  so is the underlying monotonic DL and, possibly,
 computationally effective.

 A  simple but
powerful nonmonotonic extension for
defeasible reasoning in Description Logics is proposed in
\cite{FI2009,AIJ,AIJ15}. In this approach ``typical''
or ``normal'' properties can be directly specified  by means of a
``typicality'' operator $\tip$ enriching  the underlying DL. The semantics of the $\tip$ operator is  characterized by the
core properties of nonmonotonic reasoning  axiomatized by
either \emph{preferential logic} \cite{KrausLehmannMagidor:90} or \emph{rational logic} \cite{whatdoes}. 
KLM axiomatic systems provide a  terse and well-established analysis of the core properties of
nonmonotonic reasoning.

The operator $\tip$ provides a
 natural way of expressing prototypical properties, and its
 intended meaning  is that
for any concept $C$, $\tip(C)$ singles out the instances of  $C$
that are considered as ``typical'' or ``normal''. Thus an
assertion as

\begin{quote}
``normally, a student is a young person''
\end{quote}

\noindent is
represented by

\begin{quote}
$\tip (\mathit{Student}) \sqset \mathit{YoungPerson}$
\end{quote}

\noindent We assume that a KB comprises, in addition to the standard TBox and ABox, a set of assertions of the
type $\tip(C) \sqset D$ where $D$ is a concept not mentioning
$\tip$.  For
instance, let the TBox contain:

\begin{quote}
$\mathit{SmartWorker} \sqset \mathit{Worker}$ \\
$\tip (\mathit{Worker}) \sqset \mathit{ReachableAtOffice}$\\
$\tip (\mathit{SmartWorker}) \sqset \lnot \mathit{ReachableAtOffice}$
\end{quote}

\noindent corresponding to the assertions: smart workers are workers (and this is a standard inclusion, not admitting exceptions), a typical worker is reachable at his/her office, whereas, normally, smart workers are not, since they often work at home and are reachable at their private addresses.
Suppose  further that the ABox contains
-- alternatively -- the following facts about $\mathit{paola}$:

\begin{quote}
1. $\mathit{Worker}(\mathit{paola})$ \\
2. $\mathit{Worker}(\mathit{paola}), \mathit{SmartWorker}(\mathit{paola})$ 
\end{quote}

\noindent From the different combinations of TBox and one of the above ABox assertions (either $1$ or $2$),  we would like to infer  the expected (defeasible) conclusions
about $\mathit{paola}$. These are, respectively:

\begin{quote}
1. $\mathit{ReachableAtOffice}(\mathit{paola})$\\
2. $\nott \mathit{ReachableAtOffice}(\mathit{paola})$ 
\end{quote}

\noindent Moreover, we would also like to infer (defeasible) properties of individuals
implicitly introduced by existential restrictions, for instance,
if the ABox contains

\begin{quote}
$\exists \mathit{HasColleague}.(\mathit{SmartWorker)(fabrizio)}$
\end{quote}

\noindent we would like to infer that:
\begin{quote}
$\exists \mathit{HasColleague.(\nott ReachableAtOffice)(fabrizio)}$
\end{quote}

\noindent Finally, adding irrelevant information should not affect the conclusions. From the TBox above, one should be able to infer as well

\begin{quote}
$\tip (\mathit{Worker} \mint \mathit{Slim}) \sqset \mathit{ReachableAtOffice}$\\
$\tip (\mathit{SmartWorker} \mint \mathit{Slim}) \sqset \lnot \mathit{ReachableAtOffice}$
\end{quote}

\noindent as  $\mathit{Slim}$ is irrelevant with respect to being reachable at office  or not. For the same reason, the conclusion about $\mathit{paola}$ being an instance of $\mathit{ReachableAtOffice}$
or not should not be influenced by adding
$\mathit{Slim}(\mathit{paola})$ to the ABox.

  The operator
$\tip$ is characterized by a set of postulates that are
essentially a reformulation of KLM 
axioms of preferential logic \Pe \ or rational logic \Ra, namely the assertion $\tip (C)
\sqset D$ is equivalent to the conditional assertion $C \ent D$. The operator $\tip$ is nonmonotonic, in the sense that from
$C \sqsubseteq D$ ($C$ is subsumed by $D$) we cannot infer that
$\tip(C)$ is subsumed by $\tip(D)$: even if $C \sqsubseteq D$,
the elements in $\tip(C)$ and in $\tip(D)$ can have different properties, and we can
consistently say that for some $P$, $\tip(C) \sqsubseteq P$
whereas $\tip(D) \sqsubseteq \neg P$. Models of standard DLs are extended by a function $f$ which selects the typical/most normal instances of any concept $C$, i.e. the extension of $\tip(C)$ is defined as $(\tip(C))^{\mathcal{I}}=f(C^{\mathcal{I}})$. The function $f$ satisfies a set of postulates that are a restatement of KLM's axioms. This allows the typicality operator to inherit well-established properties of nonmonotonic reasoning: as in the example above, the property known as \emph{specificity}, namely the choice of according preference to more specific information in case of conflicts among inherited properties, results ``built-in'' in the approach. 
The semantics of the $\tip$ operator can be equivalently formulated in terms of \emph{preferential} \cite{lpar2007} or
\emph{rational models} \cite{sudafricaniKR}, 
where standard DL models are extended by
an irreflexive, transitive, well-founded (and, for the rational case, modular) relation $<$ among domain elements. In this respect, $x < y$ means that $x$ is ``more normal'' than $y$, and that the typical members of a concept $C$ are the minimal elements of $C$ with respect to this relation. An element $x$ is a {\em typical instance} of some concept $C$ if it belongs to the extension of $C$ and it there is not another $C$ element $y$ such that $y < x$.
Even if the typicality operator is nonmonotonic, the resulting logic is monotonic, i.e. if a formula is entailed from a knowledge base $K$, then it is also entailed from any $K' \supseteq K$. As a consequence, the formalism is not sufficient to perform inheritance reasoning of the
kind described above. Indeed, in order to derive the expected
conclusion that Paola is not reachable at office from the above knowledge base and the fact 
$\mathit{SmartWorker}(\mathit{paola})$, we should know that
$\mathit{paola}$ is a \emph{typical} smart worker, but we do not have this information. Similarly, in
order to derive that also a typical slim worker is reachable at office, we must be able to infer or
assume that a  ``typical slim worker'' is also a
``typical worker'', since there is no reason why
it should not be the case; this cannot be derived by the logic
itself given the nonmonotonic nature  of $\tip$. The basic
monotonic logic, called $\alct$, is then too weak to enforce these extra
assumptions, therefore in \cite{AIJ} 
 the semantics of $\alct$ has been strengthened by a
minimal model semantics, which is similar, in spirit, to circumscription:
intuitively, the idea is to restrict our
consideration to models that minimize exceptions in models. 
%{\color{red} To this aim, we first define a preference relation 
%\lgcomm{Taglierei} among models, where intuitively a model $\emme_1$ is preferred to a model $\emme_2$, written $\emme_1 \preceq \emme_2$, if $\emme_1$ contains less exceptional individuals with respect to $\emme_2$; then, we define a nonmonotonic entailment restricted to models that are minimal with respect to the relation $\preceq$.}
The resulting nonmonotonic logic is called $\alctmin$, and we denote by
$\models_{min}^{\ellet}$ semantic entailment determined by minimal
models.
  It turns out that the minimal model semantics of the logic $\alctr$ corresponds to a natural extension of the notion of \emph{rational closure}, defined in \cite{whatdoes} for propositional logic, to Description Logics. This logic  presents also good computational properties: indeed, in \cite{AIJ15}, it is shown that the problem of nonmonotonic entailment is in \textsc{ExpTime}, namely reasoning about typicality and defeasible inheritance with exceptions  remains  in the same complexity class of the underlying standard Description Logic $\mathcal{ALC}$, which is already
  \textsc{ExpTime}-complete \cite{handbook}.

Preferential Description Logics have been applied to the basic system $\alc$ \cite{FI2009,AIJ,AIJ15}, as well as  to \textit{lightweight} Description Logics 
\cite{lpnmr2009,ijcai2011,LauraDTD,CasiniStracciaM19} and to more expressive DLs \cite{shiq}, witnessing the feasibility of the typicality extension.

In this chapter we first recall the nonmonotonic Description Logic $\alctr$ and its preferential semantics. Then, we will describe recent extensions of such a Description Logic: probabilistic extension, based on the DISPONTE semantics of \cite{riguzzi,disponteijcai}, suitable for formalizing the concept combination, as well as other extensions of the rational closure semantics and construction, which allow to strengthen rational closure construction and deal with the well known problem called by Pearl  ``the blocking of property inheritance problem" \cite{PearlTARK90}.
%an extension based on a semantics with multiple preference relations whence with multiple ``typicality'' operators, that allow one to distinguish different aspects of typicality/exceptionality.

\section{Description Logics of Typicality}
\vspace{-0.3cm}
In this section we present the logic $\alctr$. As mentioned in the Introduction, we allow concepts of the form $\tip(C)$,  whose intuitive meaning is that
$\tip(C)$ selects the {\em typical} instances of a concept $C$. We can therefore distinguish between the properties that
hold for all instances of concept $C$ ($C \sqsubseteq D$), and those that only hold for the typical
such instances ($\tip(C) \sqsubseteq D$).
%that we call  \tip-inclusions, where $C$ is a concept not mentioning $\tip$.
\hide{Formally, the language of  $\alctr$ is defined as follows. }
 
\vspace{-0.15cm}
 \begin{definition}\label{defelt}
 We consider an alphabet of concept names $\mathcal{C}$, of role names
$\mathcal{R}$, and of individual constants $\mathcal{O}$.
Given $A \in \mathcal{C}$ and $R \in \mathcal{R}$, we define:
\begin{quote}
 $C_R:= A \tc \top \tc \bot \tc  \nott C_R \tc C_R \sqcap C_R \tc C_R \sqcup C_R \tc \forall R.C_R \tc \exists R.C_R$\\
   $C_L:= C_R \tc  \tip(C_R)$
 \end{quote}
    A knowledge base $K$ is a pair (TBox, ABox). TBox contains a finite set
of  concept inclusions  $C_L \sqsubseteq C_R$. ABox
contains assertions of the form $C_L(a)$ and $R(a,b)$, where $a, b \in
\mathcal{O}$.
\end{definition}

The semantics of $\alctr$ can be  formulated in terms of
rational models:
 ordinary models of $\alc$ are equipped with a $\mathit{preference \ relation}$ $<$ on
the domain, whose intuitive meaning is to compare the ``typicality''
of domain elements, that is to say $x < y$ means that $x$ is more typical than
$y$. Typical members of a concept $C$, that is members of
$\tip(C)$, are the members $x$ of $C$ that are minimal with respect
to this preference relation (s.t. there is no other member of $C$
more typical than $x$).

\begin{definition}[Semantics of $\alctr$]\label{semalctr} A model $\emme$ of $\alctr$ is any
structure $\langle \Delta^\II, <, .^\II \rangle$ where: 
\begin{itemize}
\item $\Delta^\II$ is the
domain;   
\item $<$ is an irreflexive, transitive and modular (if
$x < y$ then either $x < z$ or $z < y$) relation over
$\Delta^\II$; 
\item $.^\II$ is the extension function that maps each
concept $C$ to $C^\II \subseteq \Delta^\II$, and each role $R$
to  $R^\II \subseteq \Delta^\II \times \Delta^\II$. For concepts of
$\alc$, $C^\II$ is defined in the usual way. For the $\tip$ operator, we have
$$(\tip(C))^\II = Min_<(C^\II),$$  where
$Min_<(S)= \{u: u \in S$ and $\nexists z \in S$ s.t. $z < u \}$.
\end{itemize}
Furthermore, $<$ is well-founded, i.e., there are no infinite $<$-descending chains. 
%Furthermore, $<$  satisfies the $\mathit{Well-Foundedness \ Condition}$, i.e., for all  $S \subseteq \Delta$, for all $x \in S$, either $x \in Min_<(S)$ or $\exists y \in  Min_<(S)$ such that $y < x$. 
\end{definition}

%The semantics of the $\tip$ operator can be equivalently formulated in terms of \emph{rational models} \cite{AIJ2014}: a model $\emme$ is any structure $\langle \Delta^\II, <, .^\II \rangle$ where $\Delta^\II$ is the domain, 
% $<$ is an irreflexive, transitive, well-founded and modular (for all $x, y, z$ in $\Delta^\II$, if
%$x < y$ then either $x < z$ or $z < y$) relation over $\Delta^\II$. In this respect, 
Informally, $x < y$ means that $x$ is ``more normal'' than $y$, and that the typical members of a concept $C$ are the minimal elements of $C$ with respect to this relation. An element $x
\in \Delta^\II$ is a {\em typical instance} of some concept $C$ if $x \in
C^\II$ and there is no $C$-element in $\Delta^\II$ {\em more typical} than
$x$. Elements in $\Delta^\II$ are then organized in different ``levels'' or ``ranks'' by the modularity of $<$, where all elements with rank $i$ are incomparable with each other (i.e., for all $x,y$ having rank $i$ neither $x<y$ nor $y<x$) and they are more normal than all elements with an higher rank $j>i$. Therefore, minimal $C$-elements are those having the least rank among $C$ elements. % In detail,  $.^\II$ is the extension function that maps each
%concept $C$ to $C^\II \subseteq \Delta^\II$, and each role $R$
%to  $R^\II \subseteq \Delta^\II \times \Delta^\II$. For concepts of
%$\alc$, $C^\II$ is defined as usual.  For the $\tip$ operator, given $Min_<(C^\II)=\{x \in C^\II \mid \not\exists y \in C^\II \ \mbox{s.t.}  \ y<x\}$, we have $$(\tip(C))^\II = Min_<(C^\II).$$
%  A model $\emme$  can be equivalently defined by postulating the existence of
% a function $k_{\emme}: \Delta^\II \longmapsto \mathbb{N}$, where $k_{\emme}$ assigns a finite rank to each domain element: 
% % and then letting  $x < y$ if and only if $k(x) < k(y)$. 

% \begin{definition}[Rank of a domain element]\label{definition_rank}
% Given a model $\mathcal{M}=$$\langle \Delta^\II, <, .^\II\rangle$, the rank $k_{\mathcal{M}}$  of a domain element $x \in \Delta^\II$, is the
% length of the longest chain $x_0 < \dots < x$ from $x$
% to a minimal $x_0$ (i.e. such that there is no ${x'}$ such that  ${x'} < x_0$).
% \end{definition}

% \noindent the rank function $k_{\mathcal{M}}$ and $<$ can be defined from each other  by letting $x < y$ if and only if $k_{\mathcal{M}}(x) < k_{\mathcal{M}}(y)$.

The semantics of the $\tip$ operator can also be defined by means of a set of postulates that are  a reformulation of axioms and rules of nonmonotonic entailment in rational logic {\bf R} \ \cite{whatdoes}: in this respect an assertion of the form $\tip(C) \sqsubseteq D$ is equivalent to the conditional assertion $C \mathrel{{\scriptstyle\mid\!\sim}} D$ in {\bf R}. The basic ideas are as follows: given a domain $\Delta^\II$  and an evaluation function $.^\II$, one can define a function $f_\tip : Pow(\Delta^\II) \longmapsto Pow(\Delta^\II)$
that selects  the {\em typical} instances of any $S \subseteq \Delta^\II$; in
case $S = C^\II$ for a concept $C$, the selection function selects the typical instances
of $C$, namely: $$(\tip(C))^\II = f_\tip(C^\II).$$
$f_\tip$ has the following properties for all subsets $S$ of
$\Delta^\II$, that are essentially a restatement of the properties characterizing rational logic {\bf R}:

\vspace{0.2cm}
\noindent $(f_\tip-1) \ f_\tip(S) \subseteq S$ \\
\noindent $(f_\tip-2) \ \mbox{if} \ S \not=\emptyset \mbox{, then also} \ f_\tip(S) \not=\emptyset$ \\
\noindent  $(f_\tip-3) \ \mbox{if} \ f_\tip(S) \subseteq R, \mbox{then} \ f_\tip(S)=f_\tip(S \cap R)$ \\
\noindent  $(f_\tip-4) \ f_\tip(\bigcup S_i) \subseteq \bigcup f_\tip (S_i)$\\
\noindent  $(f_\tip-5) \ \bigcap f_\tip(S_i) \subseteq f_\tip(\bigcup S_i)$\\ 
\noindent  $(f_\tip-6) \ \mbox{if} \ f_\tip(S) \cap R \not=\emptyset, \mbox{then} \ f_\tip(S \cap R) \subseteq f_\tip(S)$
\vspace{0.2cm}

\noindent
Such properties are strongly related with KLM postulates \cite{KrausLehmannMagidor:90,whatdoes} and correspond to the properties of rational consequence relations.

 Given standard definitions of satisfiability of a KB in a model, we define a notion of entailment in $\alct$. Given a query $F$ (either an inclusion $C \sqsubseteq D$ or an assertion $C(a)$ or an assertion of the form $R(a,b)$), we say that $F$ is entailed from a KB if $F$ holds in all $\alct$ models satisfying KB.

Even if
the typicality operator $\tip$ itself  is nonmonotonic (i.e.
$\tip(C) \sqsubseteq E$ does not imply $\tip(C \sqcap D)
\sqsubseteq E$), what is inferred
from a KB can still be inferred from any KB' with KB $\subseteq$
KB', i.e. the logic $\alct$ is monotonic. In order to perform useful nonmonotonic inferences,   in \cite{FI2009} and 
\cite{AIJ15} the authors have strengthened  the above semantics by
restricting entailment to a class of minimal 
models in two different ways. Intuitively, in both cases, the idea is to
restrict entailment to models that minimize the atypicality and, in the second case,
that \emph{minimize the atypical instances of a concept}. In this last case,
the resulting logic  corresponds to a notion of \emph{rational closure} built on top of $\alct$. Such a notion is a natural extension of the rational closure construction developed  by Lehmann and Magidor \cite{whatdoes} for the propositional calculus.

The nonmonotonic semantics of $\alctm$ relies on minimal rational models  that
minimize the \emph{rank  of domain elements}. Informally, given two models of
KB, one in which a given domain element $x$ has rank 2 (because for instance
$z < y < x)$, and another in which it has rank 1 (because only
$z < x$), we prefer the latter, 
as in this model the element $x$ is assumed to be ``more typical'' than in the former.

%\marginnote{\hl{We have added the formal definition for minimal models as well as for the rank of a domain element.}}

\begin{definition}[Rank of a domain element $k_{\emme}(x)$]\label{definition_rank}
Given a model $\emme=$$\langle \Delta^\II, <, .^\II\rangle$, the rank $k_{\emme}$  of a domain element $x \in \Delta^\II$, is the
length of the longest chain $x_0 < \dots < x$ from $x$
to a minimal $x_0$ (i.e. such that there is no ${x'}$ such that  ${x'} < x_0$).
\end{definition}

\begin{definition}[Minimal models]\label{Preference between models in case of fixed valuation} 
Given $\emme_1 = \langle \Delta^{\II_1}, <_1, .^{\II_1} \rangle$ and $\emme_2 =
\langle \Delta^{\II_2}, <_2, .^{\II_2} \rangle$ we say that $\emme_1$ is preferred to
$\emme_2$ if: 1. $\Delta^{\II_1} = \Delta^{\II_2}$;
2. $C^{\II_1} = C^{\II_2}$ for all concepts $C$;
3. for all $x \in \Delta^{\II_1}$, it holds that $ k_{\emme_1}(x) \leq k_{\emme_2}(x)$ and
there exists $y \in \Delta^{\II_1}$ such that $ k_{\emme_1}(y) < k_{\emme_2}(y)$.
Given a knowledge base $K$, we say that 
$\emme$ is a minimal model of $K$ if it is a model satisfying $K$ and  there is no model
$\emme'$ of $K$ satisfying it such that $\emme'$ is preferred to $\emme$.
\end{definition}

 Query entailment is then restricted to minimal {\em canonical models}. The intuition is that a canonical model contains all the individuals that enjoy properties that are consistent with the KB. A model $\emme$ is a minimal canonical model of $K$
if it satisfies $K$, it is canonical and it is  minimal among the canonical models of $K$ \footnote{In Theorem 10 in \cite{AIJ15} the authors have shown that for any consistent KB $K$ there exists a finite minimal canonical model $K$.}. A query $F$ is minimally entailed from a KB if it holds in all minimal canonical models of KB.

 In \cite{AIJ15} it is shown that  query entailment in $\alctm$ is in \textsc{ExpTime}.

\section{A Probabilistic Extension of Preferential Description Logics}
\vspace{-0.3cm}
The logic $\alctm$ imposes to consider \emph{all}  typicality assumptions that are consistent with a given KB, but this seems to be too strong in several application domains.
 It could be useful to reason about scenarios with \emph{exceptional individuals}, or one could need to assign  different \emph{probabilities} to typicality inclusions. As an  example, one could need to represent that the properties of loving sport and being active on social medias are both typical properties of students, however it could be needed to also describe that the probability of finding exceptional students not using  social networks is lower than the one of finding exceptional students not loving sport.

In \cite{ijar} an extension of preferential Description Logic of typicality called $\miaprob$ is introduced: in this logic,  typicality inclusions are equipped by \emph{probabilities of exceptionality} of the form $$\tip(C) \sqsubseteq_p D,$$ where $p \in (0,1)$.
The intuitive meaning is that
``normally, $C$s are $D$s and the probability of  having  exceptional $C$s -- not being $D$s -- is $1-p$''.
In other words, all the typical instances of the concept $C$ are also instances of the concept $D$, and the probability that a $C$ element is not also a $D$ element, i.e. it is an exceptional $C$ element, is $1-p$.
For instance, we can have 
\begin{quote}
  $ \tip(\mathit{Student}) \sqsubseteq_{0.6} \mathit{SportLover} $ \\
  $ \tip(\mathit{Student}) \sqsubseteq_{0.9} \mathit{SocialNetworkUser}  $
\end{quote}
\noindent whose intuitive meaning is that being sport lovers and social network users are both typical properties of  students, however the probability of having exceptional students not loving sport is higher than the one of finding students not using social networks, in particular we have the evidence that the probability of  having  exceptions is $40\%$ and $10\%$, respectively. 
As a difference with DLs under the distributed semantics introduced in \cite{disponteijcai,riguzzi}, where probabilistic axioms of the form $p \ :: C \sqsubseteq D$ are used to capture uncertainty in order  to represent that $C$s are $D$s with probability $p$, in the logic $\miaprob$ it is possible to ascribe typical properties to concepts and to reason about probabilities of exceptions to those typicalities.
We define different extensions of an ABox containing only some of the ``plausible'' typicality assertions: each extension represents a  scenario having a specific probability.  Then,  a notion of nonmonotonic entailment restricted to extensions whose probabilities belong to a given and fixed range is defined, in order to reason about scenarios that are not necessarily the most probable. 

Let us now recall the main formal definitions of the logic $\miaprob$.

Given a KB, we define the finite set $\CC$ of
concepts occurring in the scope of the typicality operator, i.e. $\CC= \{C \tc \tip(C) \sqsubseteq_p D \in \ \mbox{KB}\}$.
Given an individual $a$ explicitly named in the ABox, we define the set of typicality assumptions $\tip(C)(a)$ that can be minimally entailed from KB in the nonmonotonic logic $\alctm$, with $C \in \CC$.   We then consider an ordered set $\CC_{\ABox}$ of pairs $(a,C)$ of all possible assumptions $\tip(C)(a)$, for all concepts $C \in \CC$ and all individual constants $a$  in the ABox.

\begin{definition}[Assumptions in $\miaprob$]\label{cabox}
Given an $\miaprob$ KB=($\mathcal{T}, \ABox$), let $\mathcal{T}'$ be the set of inclusions of $ \mathcal{T}$ without probabilities, namely $$\mathcal{T}'=\{ \tip(C) \sqsubseteq D \tc \tip(C) \sqsubseteq_p D \in \mathcal{T}\} \ \unione \ \{C \sqsubseteq D \in \mathcal{T}\}.$$ Given a finite set of concepts $\CC$, we define, for each individual name $a$ occurring in $\ABox$:
$$\CC_a=\{ C \in \CC \tc (\mathcal{T}', \ABox) \models_{\alctm} \tip(C)(a)\}.$$
We also define 
  $\CC_{\ABox}=\{(a,C) \tc C \in \CC_a \ \mbox{and} \ a \ \mbox{occurs in} \ \ABox\}$
and we impose an order on its elements: $ \CC_{\ABox}= [(a_1,C_1), (a_2,C_2), \dots, (a_n,C_n)].$
Furthermore, we define the \emph{ordered multiset}
  $\PP_{\ABox}=[ p_1, p_2, \dots, p_n ]$, respecting the order imposed on $ \CC_{\ABox}$, where 
  \begin{center}
  $p_i= \prod\limits_{j=1}^{m} p_{ij}$ for all $\tip(C_i) \sqsubseteq_{p_{i1}} D_1, \tip(C_i) \sqsubseteq_{p_{i2}} D_2, \dots, \tip(C_i) \sqsubseteq_{p_{im}} D_m$ in $\mathcal{T}.$
  \end{center}
\end{definition}

\noindent  The ordered multiset $\PP_{\ABox}$ is a tuple of the form $[ p_1, p_2, \dots, p_n ]$, where $p_i$ is the probability
 of the assumption $\tip(C)(a)$, such that $(a,C) \in  \CC_{\ABox}$ at position $i$. $p_i$ is  the product of all the probabilities $p_{ij}$ of typicality inclusions $\tip(C) \sqsubseteq_{p_{ij}} D$ in the TBox. 

We consider different extensions $\widetilde{\ABox_i}$ of the ABox and we equip them with a probability $\pgrande_i$. 
Starting from  $\PP_{\ABox}=[p_1, p_2, \dots, p_n]$, the first step is to build all alternative tuples where $0$ is used in place of some $p_i$ to represent that the corresponding typicality assertion $\tip(C)(a)$ is no longer assumed (Definition \ref{possibili estensioni}). Furthermore, we define the \emph{extension} of the ABox corresponding to a string so obtained (Definition \ref{corrispondenza d abox}). In this way, the highest probability is assigned to the extension of the ABox corresponding to $\PP_{\ABox}$, where all typicality assumptions are considered. The probability decreases in the other extensions, where some typicality assumptions are discarded, thus $0$ is used in place of the corresponding $p_i$. The probability of an extension $\widetilde{\ABox_i}$ corresponding to a string $\PP_{\ABox_i}=[p_{i1}, p_{i2}, \dots, p_{in}]$ is defined as the product of probabilities $p_{ij}$ when $p_{ij} \diverso 0$, i.e. the probability of the corresponding typicality assumption when this is selected for the extension, and $1-p_{j}$ when $p_{ij} =0$, i.e. the corresponding typicality assumption is discarded, that is to say the extension contains an exception to the inclusion.

% Let us now introduce formal definitions for the above mentioned notions of string of  assumptions and of extension of an ABox corresponding to a string.

\begin{definition}[Strings of possible assumptions $\esse$]\label{possibili estensioni}
Given a KB=($\TBox, \ABox$), let the set $ \CC_{\ABox}$ and  $\PP_{\ABox}=[p_1, p_2, \dots, p_n]$ be as in Definition \ref{cabox}.
We define the set $\esse$ of all the \emph{strings of possible assumptions} with respect to KB as 
\begin{center}
$\esse=\{[s_1, s_2, \dots, s_n] \tc$  $\forall i=1, 2, \dots, n \ \mbox{either} \ s_i=p_i \ \mbox{or} \ s_i=0\}$
\end{center}
\end{definition}

\begin{definition}[Extension of  ABox]\label{corrispondenza d abox}
Let  KB=($\TBox, \ABox$),   $\PP_{\ABox}=$ $[p1, p_2, \dots, p_n]$  and  $ \CC_{\ABox}=[(a_1, C_1), (a_2, C_2), \dots, (a_n, C_{n})]$ as in Definition \ref{cabox}.
Given a string of possible assumptions $[s_1, s_2, \dots, s_n] \in \esse$ of Definition \ref{possibili estensioni}, we define the extension 
$\widetilde{\ABox}$ of $\ABox$  with respect to  $\CC_{\ABox}$ and $\esse$ as:
\[
   \widetilde{\ABox}=\{ \tip(C_i)(a_i) \tc (a_i, C_i) \in  \CC_{\ABox} \ \mbox{and} \ s_i \diverso 0 \}
\]
We also define the probability  of $\widetilde{\ABox}$ as $\pgrande_{\widetilde{\ABox}}=\prod\limits_{i=1}^{n} \chi_i $ where 
$\chi_i  =  
\bigg \{
  \begin{tabular}{lll}
  $p_i$ & \qquad & if \ $s_i \diverso 0$ \\
  $1-p_i$ & \qquad & if \ $s_i = 0$
  \end{tabular}
$
 \end{definition}

\noindent It can be observed that, in $\alctm$,  the set of typicality assumptions that can be inferred from a KB corresponds to the extension of $\ABox$ corresponding to the string $\PP_{\ABox}$ (no element  is set to $0$): all the typicality assertions of individuals occurring in the ABox, that are consistent with the KB, are assumed. On the contrary, in $\alct$, no typicality assumptions can be derived from a KB, and this corresponds to extending $\ABox$ by the assertions corresponding to the string $[0, 0, \dots, 0]$, i.e. by the empty set. It is easy to observe that we obtain a probability distribution over  extensions of  $\ABox$. 

We have now all the ingredients for recalling formal definitions for nonmonotonic entailment in the Description Logic $\miaprob$. Intuitively, given KB and a query $F$, we distinguish two cases: 
\begin{itemize}
\item if $F$ is an inclusion $C \sqsubseteq D$, then it is entailed from KB if it is minimally entailed from KB' in the logic $\alctmin$, where KB' is obtained from KB by removing probabilities of exceptions, i.e. by replacing each typicality inclusion $\tip(C) \sqsubseteq_p D$ with $\tip(C) \sqsubseteq D$;
   \item if $F$ is an ABox fact $C(a)$, then it is entailed from KB if it 
   is entailed in the monotonic $\alct$ from the knowledge bases including the extensions of the ABox of Definition \ref{corrispondenza d abox}.  
 \end{itemize}  
  We provide both (i) a notion of entailment restricted to scenarios whose probabilities belong to a given range and (ii) a notion of probability of the entailment of a query $C(a)$, as the sum of the probabilities of all extensions from which $C(a)$ is so entailed.

Given a knowledge base KB and two real numbers $p$ and $q$, we write KB $\models^{\sx p, q \dx}_{\miaprob} F$ to represent that $F$ follows -- or is entailed -- from KB restricting reasoning to scenarios whose probabilities range from $p$ to $q$. We distinguish the case in which the query is a TBox inclusion from the one in which it is an ABox assertion. 

\begin{definition}[Entailment in $\miaprob$]\label{entailment in miaprobTBox}
Given a KB=($\TBox, \ABox$), two reals $p, q \in (0,1]$, and a query $F$ which is a TBox inclusion either $C \sqsubseteq D$ or $\tip(C) \sqsubseteq D$, we say that $F$ is  entailed from KB in $\miaprob$ in range $\sx p, q \dx$, written 
KB $\models^{\sx p, q \dx}_{\miaprob} F$, if ($\TBox', \ABox$) $\models_{\alctmin} F$, where $\TBox'=\{\tip(C) \sqsubseteq D \tc \tip(C) \sqsubseteq_r D \in \TBox\} \ \unione \ \{C \sqsubseteq D \in \TBox\}$.
\end{definition}

\begin{definition}[Entailment in $\miaprob$]\label{entailment in miaprobABox}
Given a KB=($\TBox, \ABox$), given $\CC$ a set of concepts, and given $p, q \in (0,1]$, let $\mathcal{E}=\{\widetilde{A_1}, \widetilde{A_2}, \dots, \widetilde{A_k}\}$ be the set of  extensions of $\ABox$ of Definition \ref{corrispondenza d abox} with respect to $\CC$, whose probabilities are such that $p \leq \pgrande_1 \leq q, p \leq \pgrande_2 \leq q, \dots, p \leq \pgrande_k \leq q$. Let $\TBox'=\{\tip(C) \sqsubseteq D \tc \tip(C) \sqsubseteq_r D \in \TBox\} \ \unione \ \{C \sqsubseteq D \in \TBox\}$. Given a query $F$ which is an ABox assertion $C(a)$, where $a \in \mathcal{O}$, we say that $F$ is  entailed from KB in $\miaprob$ in range $\sx p, q \dx$, written KB $\models^{\sx p, q \dx}_{\miaprob} F$,  if ($\TBox', \ABox  \  \unione \ \widetilde{\ABox_i}$) $\models_{\alct} F$ for all $\widetilde{\ABox_i} \in \mathcal{E}$. \\
We also define the probability of the entailment of a query as $\pgrande(F)=\sum\limits_{i=1}^{k} \pgrande_{i}$.
\end{definition}

\noindent It is worth noticing that, in Definition \ref{entailment in miaprobTBox}, probabilities $p$ and $q$ do not play any role: indeed, probabilities of scenarios are related to ABox extensions, that are not involved when we are reasoning about TBoxes. As already mentioned, in this case entailment in $\miaprob$ corresponds to entailment in the nonmonotonic Description Logic $\alctmin$.
In \cite{ijar} it is shown that the problem of entailment in the logic $\miaprob$ is \textsc{ExpTime} complete.

\section{A Logic for Concept Combination} \label{logica}
\vspace{-0.3cm}
In this section we exploit preferential Description Logics in order to tackle the problem of combining two prototypical descriptions. This problem is important because represents a classical issue in formal and cognitive semantics. 
%In fact, the invention of novel concepts obtained by combining the typical knowledge of pre-existing ones, is among the most creative cognitive abilities exhibited by humans. 
This generative phenomenon, indeed, highlights some crucial aspects of the knowledge processing capabilities in human cognition and concerns high-level capacities associated to creative thinking and problem solving. Still, it represents an open challenge in the field of Artificial Intelligence (AI) \cite{colton2012computational}. 
Dealing with this problem requires, from an AI perspective, the harmonization of two conflicting requirements that are hardly accommodated in formal ontologies \cite{frixione2011representing}: the need of a syntactic and semantic compositionality and that one concerning the exhibition of typicality effects.  
According to a well-known argument, in fact, prototypes are not compositional. The argument runs as follows: consider a concept like \emph{pet fish}. It results from the composition of the concept \emph{pet} and of the concept \emph{fish}. However, the prototype of \emph{pet fish} cannot result from the composition of the prototypes of a pet and a fish: e.g. a typical pet is furry and warm, a typical fish is grayish, but a typical pet fish is neither furry and warm nor grayish (typically, it is red). 

In \cite{ismis2018,jetai} we have provided a framework able to account for this type of human-like concept combination by introducing a probabilistic extension of a Description Logic of typicality called $\pfl$ (typicality-based compositional logic). 
The nonmonotonic Description Logic $\pfl$ combines the semantics based on the rational closure of $\alctr$ with the probabilistic DISPONTE semantics \cite{disponteijcai}.
By taking inspiration from \cite{lieto2017dual}, we consider two types of properties associated to a given concept: rigid and typical. Rigid properties are those that hold under any circumstance, e.g. $C \sqsubseteq D$ (all $C$s are $D$s). Typical properties are represented by inclusions equipped by a probability. Additionally,  we employ a cognitive heuristic for the identification of a dominance effect between the concepts to be combined, distinguishing between HEAD and MODIFIER. 

The language of $\pfl$ extends the basic DL $\mathcal{ALC}$ by \emph{typicality inclusions} of the form $\tip(C) \sqsubseteq D$ equipped by a real number $p \in (0.5,1)$ representing its probability, whose  meaning is that ``normally, $C$s are also $D$ with probability $p$''.

\begin{definition}[Language of $\pfl$]
We consider an alphabet $\mathtt{C}$ of concept names, $\mathtt{R}$ of role names, and $\mathtt{O}$ of individual constants.
Given $A \in \mathtt{C}$ and $R \in \mathtt{R}$, we define:

\vspace{0.2cm}

 $ C, D:= A \mid \top \mid \bot \mid  \lnot  C \mid  C \sqcap  C \mid  C \sqcup  C \mid \forall R. C \mid \exists R. C$
 
\vspace{0.2cm}

\noindent   We define a knowledge base $\kk=\langle \RR, \TT, \AAA \rangle$ where:
\vspace{-0.13cm}
\begin{itemize}
\item $\RR$ is a finite set of rigid properties of the form $C \sqsubseteq D$; 
\item $\TT$ is a finite set of typicality properties of the form $p \ :: \ \tip(C) \sqsubseteq D$, where $p \in (0.5,1) \subseteq \mathbb{R}$ is the probability of the inclusion;
 \item $\AAA$ is the ABox, i.e. a finite set of formulas of the form either $C(a)$ or $R(a,b)$, where $a,b \in \mathtt{O}$.
\end{itemize}
\end{definition}

\noindent It is worth noticing that we avoid typicality inclusions with degree 1. Indeed, an inclusion $1 \ :: \ \tip(C) \sqsubseteq D$ would mean that it is a certain property, that we represent with $C \sqsubseteq D \in \RR$. Also, observe that we only allow typicality inclusions equipped with probabilities $p > 0.5$. Indeed, the very notion of typicality derives from the one of probability distribution, in particular typical properties attributed to entities are those characterizing the majority of instances involved. Moreover, in our effort of integrating two different semantics -- DISPONTE and typicality logic -- the choice of having probabilities higher than $0.5$ for typicality inclusions seems to be the only compliant with both formalisms. In fact, despite the DISPONTE semantics allows to assign also low probabilities/degrees of belief to standard inclusions, in the logic $\pfl$ it would be misleading to also allow low probabilities for typicality inclusions. 

 Following from the DISPONTE semantics, each axiom is independent from each others. This this allows us to deal with conflicting typical properties equipped with different probabilities. 
A model $\emme$ of $\pfl$ is as in Definition \ref{semalctr}. Probabilities equipping typicality inclusions do not play any role in the models, but they are used to define \emph{scenario} of the composition of concepts in the distributed semantics. Intuitively, a scenario is a knowledge base obtained by adding to all rigid properties in $\RR$ and to all ABox facts in $\AAA$ only \emph{some} typicality properties. More in detail, we define an {\em atomic choice} on each typicality inclusion, then we define a {\em selection} as a set of atomic choices in order to select which typicality inclusions have to be considered in a scenario.

\begin{definition}[Atomic choice]
Given $\kk=\langle \RR, \TT, \AAA \rangle$, where $\TT = \{ E_1 = q_1 \ :: \tip(C_1) \sqsubseteq D_1,  \dots, E_n = q_n \ :: \tip(C_n) \sqsubseteq D_n \}$ we define ($E_i$, $k_i$) an \emph{atomic choice} for some $i \in \{1, 2, \dots, n\}$, where $k_i \in \{0, 1\}$. 
\end{definition}

\begin{definition}[Selection]\label{def:selection}
Given  $\kk=\langle \RR, \TT, \AAA \rangle$, where $\TT = \{ E_1 = q_1 \ :: \tip(C_1) \sqsubseteq D_1,  \dots, E_n = q_n \ :: \tip(C_n) \sqsubseteq D_n \}$ and a set of atomic choices $\nu$, we say that $\nu$ is a \emph{selection} if, for each $E_i$, one decision is taken, i.e. either ($E_i$, 0) $\in \nu$ and ($E_i$, 1) $\not\in \nu$ or  ($E_i$, 1) $\in \nu$ and ($E_i$, 0) $\not\in \nu$ for $i=1, 2, \dots, n$. The probability of  $\nu$ is $P(\nu) = \prod\limits_{(E_i,1) \in \nu} q_i \prod\limits_{(E_i,0) \in \nu} (1-q_i)$.
\end{definition}

\begin{definition}[Scenario]
Given  $\kk=\langle \RR, \TT, \AAA \rangle$, where $\TT = \{ E_1 = q_1 \ :: \tip(C_1) \sqsubseteq D_1, \dots, E_n = q_n \ :: \tip(C_n) \sqsubseteq D_n \}$ and given a selection $\sigma$, we define a \emph{scenario} $w_\sigma=\langle \RR,  \{E_i \mid (E_i, 1) \in \sigma\}, \AAA \rangle.$  
We also define the probability of a scenario $w_\sigma$ as the probability of the corresponding selection, i.e. $P(w_\sigma)=P(\sigma)$.
Last, we say that a scenario is \emph{consistent} when it admits a model in the logic $\pfl$.
\end{definition}

\noindent We denote with $\allworlds$ the set of all scenarios. It immediately follows that the probability of a scenario $P(w_\sigma)$ is a probability distribution over scenarios, that is to say $\sum\limits_{w \in \allworlds} P(w) = 1$.

 Given a KB $\kk=\langle \RR, \TT, \AAA \rangle$ and given two concepts $C_H$ and $C_M$ occurring in $\kk$, our logic allows to define the compound concept $C$ as the combination of the HEAD $C_H$ and the MODIFIER $C_M$, where $C \sqsubseteq C_H \sqcap C_M$ and the typical properties of the form $\tip(C) \sqsubseteq D$ to ascribe to the concept $C$ are obtained in the set of scenarios that: 1) are consistent; 2) are not trivial, i.e. those with the highest probability, in the sense that the scenarios considering \emph{all} properties that can be consistently ascribed to $C$ are discarded; 3) are those giving preference to the typical properties of the HEAD $C_H$ (with respect to those of the MODIFIER $C_M$) with the highest probability. 
 
 Notice that, in case of conflicting properties like $D$ and $\lnot D$, given two scenarios $w_1$ and $w_2$, both belonging to the set of consistent scenarios with the highest probability and such that an inclusion $p_1 \ :: \ \tip(C_H) \sqsubseteq D$ belongs to $w_1$ whereas $p_2 \ :: \ \tip(C_M) \sqsubseteq \lnot D$ belongs to $w_2$, the scenario $w_2$ is discarded in favor of $w_1$.

%\begin{enumerate}
 %  \item are consistent; 
  % \item are not trivial, i.e. those with the highest probability, in the sense that the scenarios considering \emph{all} properties that can be consistently ascribed to $C$ are discarded;
   %\item are those giving preference to the typical properties of the HEAD $C_H$ (with respect to those of the MODIFIER $C_M$) with the highest probability. Notice that, in case of conflicting properties like $D$ and $\lnot D$, given two scenarios $w_1$ and $w_2$, both belonging to the set of consistent scenarios with the highest probability and such that an inclusion $p_1 \ :: \ \tip(C_H) \sqsubseteq D$ belongs to $w_1$ whereas $p_2 \ :: \ \tip(C_M) \sqsubseteq \lnot D$ belongs to $w_2$, the scenario $w_2$ is discarded in favor of $w_1$.
%\end{enumerate}

\noindent In order to select the wanted scenarios we apply points 1, 2, and 3 above to blocks of scenarios with the same probability, in decreasing order starting from the highest one. 

\hide{More in detail, we first discard all the inconsistent scenarios, then we consider the remaining (consistent) ones in decreasing order by their probabilities. We then consider the blocks of scenarios with the same probability, and we proceed as follows:
\begin{itemize}
   \item we discard those considered as \emph{trivial}, consistently inheriting all (or most of) the properties from the starting concepts to be combined;
   \item among the remaining ones, we discard those inheriting properties from the MODIFIER in conflict with properties inherited from the HEAD in another scenario of the same block (i.e., with the same probability);
   \item if the set of scenarios of the current block is empty, i.e. all the scenarios have been discarder either because trivial or because preferring the MODIFIER, we repeat the procedure by considering the block of scenarios, all having the immediately lower probability;
   \item the set of remaining scenarios are those selected by the logic $\pfl$.
\end{itemize}

}

The knowledge base obtained as the result of combining concepts $C_H$ and $C_M$ into the compound concept $C$ is called $C$-\emph{revised} knowledge base: 
%\vspace{-0.15cm}
 $$\kk_C=\langle \RR, \TT \cup \{p \:: \ \tip(C) \sqsubseteq D\}, \AAA \rangle,$$ for all $D$ such that $\tip(C) \sqsubseteq D$ is entailed in $w$.
The probability $p$ is defined as follows: if $D$ is a property inherited either from the HEAD (or from both the HEAD and the MODIFIER),  then $p$ corresponds to the probability of such inclusion of the HEAD in the initial knowledge base, i.e. $p \:: \ \tip(C_H) \sqsubseteq D \in \TT$; otherwise, $p$ corresponds to the probability of such inclusion of a MODIFIER in the initial knowledge base, i.e. $p \:: \ \tip(C_M) \sqsubseteq D \in \TT$.
Notice that, since the $C$-\emph{revised} knowledge base is still in the language of the $\pfl$ logic, we can iteratively repeat the same procedure in order to combine not only atomic concepts, but also compound concepts.

As an example, consider the following instantiation of the above mentioned \emph{pet fish} problem: let $\kk=\langle \RR, \TT, \AAA \rangle$ be a  KB, where the ABox $\AAA$ is empty, the set of rigid inclusions is
    $\RR=\{\mathit{Fish} \sqsubseteq \forall \mathit{livesIn}.\mathit{Water}\}$
and the set of typicality properties $\TT$ is:
\begin{quote}
      (1) \ $0.8  \ :: \ \tip(\mathit{Fish}) \sqsubseteq \lnot \mathit{Affectionate}$ 
      $\qquad\quad  \ $ 
    (2) \ $0.6 \ :: \ \tip(\mathit{Fish}) \sqsubseteq \mathit{Greyish}$ \\
    (3) \ $0.9 \ :: \ \tip(\mathit{Fish}) \sqsubseteq \mathit{Scaly}$ 
          $\qquad\qquad\qquad \ \ \ \   $ 
    (4) \ $0.8 \ :: \ \tip(\mathit{Fish}) \sqsubseteq \lnot \mathit{Warm}$ \\
    (5) \ $0.9 \ :: \ \tip(\mathit{Pet}) \sqsubseteq \forall \mathit{livesIn}.(\lnot \mathit{Water})$ \\
    (6) \ $0.8  \ :: \ \tip(\mathit{Pet}) \sqsubseteq \mathit{Affectionate}$ 
    $\qquad\qquad   $
    (7) \ $0.8 \ :: \ \tip(\mathit{Pet}) \sqsubseteq \mathit{Warm}$ 
\end{quote}

\noindent In the logic $\pfl$ the not trivial scenario defining prototypical properties of a pet fish is defined from the selection $\sigma=\{(1,1), (2,0), (3,1), (4,1), (5,0), (6,0), (7,0)\}$, containing inclusions (1), (3), and (4).
%and the resulting scenario $w_\sigma$ is as follows:
%\begin{quote}
%  \begin{enumerate}
%   \item[1.] $0.8  \ :: \ \tip(\mathit{Fish}) \sqsubseteq \lnot \mathit{Affectionate}$ 
%   \item[3.] $0.9 \ :: \ \tip(\mathit{Fish}) \sqsubseteq \mathit{Scaly}$ 
%   \item[4.] $0.8 \ :: \ \tip(\mathit{Fish}) \sqsubseteq \lnot \mathit{Warm}$ 
%  \end{enumerate}
%\end{quote}
 The resulting $\mathit{Pet} \ \sqcap \ \mathit{Fish}$-revised knowledge base is $\kk_{\mathit{Pet} \ \sqcap \ \mathit{Fish}}=\langle \{\mathit{Fish} \sqsubseteq \forall \mathit{livesIn}.\mathit{Water}\}, \TT \cup \TT', \emptyset \rangle$, where $\TT$ is:
  \begin{quote}
     $0.8 \ :: \ \tip(\mathit{Pet} \sqcap \mathit{Fish}) \sqsubseteq \lnot \mathit{Affectionate}$  \\
    $0.9 \ :: \ \tip(\mathit{Pet} \sqcap \mathit{Fish}) \sqsubseteq \mathit{Scaly}$ \\
    $0.8 \ :: \ \tip(\mathit{Pet} \sqcap \mathit{Fish}) \sqsubseteq \lnot \mathit{Warm}$ 
\end{quote}

\noindent Notice that in the Description Logic $\pfl$, adding a new inclusion $\tip(\mathit{Pet} \sqcap \mathit{Fish}) \sqsubseteq \mathit{Red}$, would not be problematic: this means that our formalism is able to tackle the cognitive phenomenon of \emph{attributes emergence} for the new compound concept \cite{hampton1987inheritance}.

 In \cite{jetai,ismis2018} we have shown that reasoning in $\pfl$ in the revised knowledge is \textsc{ExpTime}-complete. The proposed logic has been adopted in concrete computational creativity applications. In particular, it has been used both in a system for the automatic, goal-directed, knowledge augmentation of dynamic knowledge bases \cite{csr19beyond} and as a logic engine for a serendipity-based recommender system, applied to the RaiPlay platform, able to generate and suggest novel narrative contents to the users \cite{ecai2020}. 
 
 \section{Refinements of the Rational Closure}
\vspace{-0.3cm}
Lehmann and Magidor's rational closure (RC) construction \cite{whatdoes} was first considered for DLs in \cite{casinistraccia2010} and 
later it was studied for $\alc$ in \cite{AIJ15,TesiMoodley2016}, % \cite{CasiniDL2013,dl2013}.
where polynomial reductions of the RC to standard DLs have been considered.
%The rational closure can be computed by exploiting polynomial reductions to standard DLs \cite{AIJ15,TesiMoodley2016},and its construction requires a quadratic number of entailments to the underlying DL reasoner. 
However,  RC suffers from a well known problem
called by Pearl \cite{PearlTARK90} ``the blocking of property inheritance problem", and by Benferhat et al. the ``drowning problem" \cite{BenferhatIJCAI93}.
The problem can be summarised as follows:
if a subclass of a class $C$ is exceptional with respect to $C$ for a given aspect, it is exceptional tout court and does not inherit any of the typical properties of $C$. 

Refinements of the RC construction, avoiding this problem, have been studied  in the context of propositional logic, among which the lexicographic closure introduced by Lehmann \cite{lex} %\cite{Lehmann95},
was later extended to DLs  by Casini and Straccia  \cite{Casinistraccia2012}.
 Besides this proposal, in the context of DLs other approaches have been considered to deal with the above mentioned problem of RC.
The same authors have developed an inheritance-based approach for defeasible DLs  \cite{CasiniJAIR2013}.
Casini et al. have introduced the notions of  basic and minimal  Relevant Closure \cite{Casini2014}  as extensions of RC, 
where relevance is based on the notion of justification. In \cite{GliozziAIIA2016} Gliozzi has defined a multipreference semantics for defeasible inclusions in which  models are equipped with several preference relations, providing a refinement of the RC semantics. Two closure constructions, the  MP-closure and the Skeptical closure,  have been proposed as weaker refinements of the rational closure for $\alc$, approximating the multipreference semantics.
The  logic of overriding ${\cal DL}^N$  \cite{bonattiAIJ15} may
exploit RC to determine specificity of defaults, and do not suffer from the above problem. In a way, when building on RC to determine the 
ranking of concepts (rather than building on the concept hierarchy), also ${\cal DL}^N$ can be regarded as a refinement of RC. 

In this section we will present the ideas underlying the 
skeptical closure construction trough some examples. This closure, as the MP-closure, can be regarded as  a weaker variant of the lexicographic closure.
The skeptical closure is weaker than the MP-closure \cite{Pruv2018,arXiv_Skeptical_closure_TR_UPO}, which is in turn weaker than the 
multipreference semantics  \cite{Ecsqaru19}. 
For a given KB and a query, the skeptical closure is based on the construction of a single base, while the MP-closure (as the lexicographic closure) requires building multiple (and, in the worst case, exponentially many) bases (sets of defeasible inclusions). 

Let us consider the following example from \cite{arXiv_Skeptical_closure_TR_UPO}.
%\begin{example} \label{exa:BabyPenguin} 
Consider a knowledge base $K=({\cal T} ,{\cal A})$, where ${\cal A}=\emptyset$ and  ${\cal T}$ contains the following inclusions: 

(1) $\mathit{Penguin \sqsubseteq Bird}$  \ \ \ \ \ \ \ \ \ \ \ \ \ \ \ \ \ \ \ \ \ \ \ \ \ \ \ \ \  \ \ \ \ \ \ \ (2) $\mathit{BabyPenguin \sqsubseteq Penguin}$
 
(3) $\mathit{\tip(Bird) \sqsubseteq  Fly}$ \ \ \ \ \ \ \ \ \ \ \ \ \ \ \ \ \ \ \ \ \ \ \ \ \ \ \ \ \ \ \ \ \ \ \ \ \ \ (4) $\mathit{\tip(Bird) \sqsubseteq  NiceFeather}$

(5) $\mathit{\tip(Penguin) \sqsubseteq \neg Fly}$  \ \ \ \ \ \ \ \ \ \ \ \ \ \ \ \ \ \ \ \ \ \ \ \ \ \ \ \ \   (6) $\mathit{\tip(Penguin) \sqsubseteq BlackFeather}$ 

(7) $\mathit{\tip(BabyPenguin)}$ $ \sqsubseteq$ $\mathit{ \neg BlackFeather}$

\noindent
Here, we expect that the defeasible property of birds having a nice feather is inherited by typical penguins,  even though penguins are exceptional birds regarding flying.
We also expect that typical baby penguins inherit the defeasible property of penguins that they do not fly,  
although the defeasible property $\mathit{BlackFeather}$ is instead overridden for typical baby penguins, and that they inherit the typical property of birds of  having nice feather.
The RC construction assigns rank $0$ to $\mathit{Bird}$, rank $1$ to $\mathit{Penguin}$, and rank $2$ to $\mathit{BabyPenguin}$, the more specific concept having the higher rank.
RC does not allow the conclusion that penguins have nice feather, as penguins are exceptional w.r.t. birds concerning flying  and, hence, they do not inherit any of the properties of birds.
Similarly, it does not allow the conclusion that typical baby penguins (being penguins) do not fly, as baby penguins are exceptional w.r.t. penguins concerning their color. Hence, baby penguins neither inherit properties of penguins nor properties of birds.

The skeptical closure addresses the problem above by building a base for a given concept $B$ (e.g., $\mathit{BabyPenguin}$), by collecting all the defeasible inclusions which are compatible with $B$ 
and adding them to the defeasible inclusions with the same rank as $B$, forming a base.
If $B$ has rank $i$, all the defeasible inclusions with rank $i$ are in the base. 
Then the construction proceeds rank by rank, from rank $i-1$ to rank $0$.
For each rank $k$, if the defeasible inclusions with rank $k$ individually compatible with $B$ (i.e. those which are not overridden by more specific inclusions with higher rank) are all together  consistent with $B$, they are all added to the base. If not, there are conflicting defaults with rank $k$ and we stop. 
In the example,  $\mathit{BabyPenguin}$ has rank 2 and inclusion (7) is added to the base (as well as strict inclusions). 
Defaults (5) and (6) are the only defeasible inclusions with rank 1.
Default (5) is  compatible with $\mathit{BabyPenguin}$, given default (7) and the strict inclusions, while default (6) is not (is overridden by (7)).
Hence, (5) is added to the base. 
(3) and (4) are defaults with rank 0. (3) is overridden by (5). (4) is added to the base. There are no conflicting defaults with the same rank.
The skeptical closure for $\mathit{BabyPenguin}$ 
then contains (in addition to strict inclusions) defeasible inclusions  (4), (5), and (7).
%but neither  inclusion (3), which is overridden by the more specific (5), nor inclusion (6), which is overridden by (7).
From this base, using entailment in $\alctr$, we conclude that typical baby penguins have nice feather and do not fly.
%\end{example}

The same conclusion can be derived by the stronger MP-closure and lexicographic closure, as in both cases there is a unique minimal basis for 
$\mathit{BabyPenguin}$, coinciding with the one above.
The same conclusions holds 
as well in ${\cal DL}^N$ and in both the basic and the minimal Relevant closures. 
Let us consider the following simple example in which the  skeptical closure is weaker than all other constructions.

(1) $\mathit{\tip(Eagle) \sqsubseteq  Fly}$ \ \ \ \ \ \ \ \ \ \ \ \ \ \ \ \ \ \ \ \ \ \ \ \ \ \ \ \ \ \ \ \ \ \ \ \ \  (2) $\mathit{\tip(Eagle) \sqsubseteq  NiceFeather}$

(3) $\mathit{\tip(OldAnimal) \sqsubseteq \neg NiceFeather}$  \ \ \ \ \ \ \ \ \ \   (4) $\mathit{OldEagle \equiv  Eagle \sqcap OldAnimal}$

\noindent
For concept  $\mathit{OldEagle}$, the defeasible inclusions (2) and (3) are conflicting, and they have all the same rank 0 (the same specificity).
In MP-closure and lexicographic closure there are two bases for $\mathit{OldEagle}$, one containing defaults (1) and (2) and the other containing defaults (1) and (3). As from both bases we can conclude that old eagles fly, then defeasible inclusion $\mathit{\tip(OldEagle) \sqsubseteq  Fly}$ holds. 
However, the skeptical closure does not allow this conclusion as all defaults $(1), (2), (3)$ are individually compatible with concept $\mathit{OldEagle}$, they have all the same rank (rank 0) and are conflicting, so that the skeptical closure discards them all.  The relevant closure in this case would behave as the MP-closure and the lexicographic closure and would accept the conclusion that normally old birds fly.
The logic of overriding  ${\cal DL}^N$ would find out that there is a conflict between the defaults (2) and (3), none of which is overridden by more specific properties making  the prototype of concept $\mathit{OldBird}$ inconsistent.

Entailment defined by the skeptical closure satisfies all KLM properties of a preferential consequence relation. 
Skeptical closure has been proved to be weaker than MP-closure, but neither weaker nor stronger than basic and  minimal Relevant closures.
We refer to \cite{arXiv_Skeptical_closure_TR_UPO} for an example in which skeptical closure is stronger than both basic and minimal Relevant closures.
In the DL case,  lexicographic closure is stronger than  basic and  minimal  Relevant closure  \cite{Casini2014}.
In the propositional case, it has been proved \cite{reconstructionMPclosure_arXiv2019} that  MP-closure is stronger than relevant closure, but
weaker than Lehmann's lexicographic closure, so that skeptical closure is also weaker than lexicographic closure in the propositional case.

\section{Conclusions}
\vspace{-0.3cm}
We have provided an overview of 
 preferential DLs of typicality,
 which allow a user to represent and reason about prototypical properties. Recently, preferential DLs have been extended in two directions: on the one hand, probabilistic extensions have been applied to the task of commonsense concept combination, on the other hand, a strengthening of the rational closure semantics has been proposed in order to avoid the well known problem of inheritance blocking. For what concerns the first extension, we aim at extending the proposed approach to more expressive DLs, such as those underlying the standard OWL language, and at implementing efficient reasoners for commonsense concept combination.
 Concerning  refinements of RC, as RC definition has been investigated for expressive DLs \cite{LauraDTD},
 for low complexity DLs \cite{lpnmr2009,ijcai2011,LauraDTD,CasiniStracciaM19}, and for all DLs \cite{Bonatti2019}, 
a natural question is whether skeptical closure  and other closure constructions can as well be extended to these DLs.

\bibliography{pozzSSW.bib}

\end{document}